%% file: main.tex
\documentclass[runningheads]{llncs}

\usepackage{eccv}

\usepackage{eccvabbrv}

\usepackage{graphicx}
\usepackage{booktabs}

\usepackage[accsupp]{axessibility}  

\usepackage{hyperref}

\usepackage{orcidlink}

\usepackage{graphicx}
\usepackage{amsmath}
\usepackage{amssymb}
\usepackage{booktabs}
\usepackage{multirow}
\usepackage{tabularx}
\usepackage{enumitem}
\usepackage{gensymb}
\usepackage{colortbl}
\usepackage{caption}
\usepackage{makecell}
\usepackage{wrapfig}
\usepackage{amssymb}
\usepackage{pifont}
\usepackage{marvosym}

\definecolor{mycyan}{cmyk}{.1,0,0,0}
\newcommand{\cmark}{\ding{51}}%
\newcommand{\cmarkg}{\textcolor{lightgray}{\ding{51}}}%
\newcommand{\xmark}{\ding{55}}%
\newcommand{\xmarkg}{\textcolor{lightgray}{\ding{55}}}%

\newcommand{\mypara}[1]{\vspace{1mm}\noindent\textbf{#1}}
\newcommand{\name}{RDA-Driver}

\definecolor{mygray}{gray}{.95}

\title{Making Large Language Models Better Planners with Reasoning-Decision Alignment}

\titlerunning{RDA-Driver: Reasoning-Decision Alignment}

\makeatletter
\newcommand{\printfnsymbol}[1]{%
  \textsuperscript{\@fnsymbol{#1}}%
}

\makeatother

\author{Zhijian Huang\inst{1}\thanks{Equal contribution.}, Tao Tang\inst{1}\printfnsymbol{1}, Shaoxiang Chen\inst{2}, Sihao Lin\inst{3}, Zequn Jie\inst{2}\textsuperscript{\Letter}, \\
Lin Ma\inst{2}, Guangrun Wang\inst{4}, 
Xiaodan Liang\inst{1,5}\textsuperscript{\Letter}
}

\authorrunning{Z. Huang et al.}

\institute{
  \textsuperscript{1}Shenzhen Campus of Sun Yat-sen University \quad \textsuperscript{2}Meituan Inc. \\ \quad 
  \textsuperscript{3}University of Technology Sydney \quad 
  \textsuperscript{4}Sun Yat-sen University \quad \\
  \textsuperscript{5}Research Institute of Multiple Agents and Embodied Intelligence, Peng Cheng Laboratory, Shenzhen, China \\
}

\begin{document}
\maketitle

\newcommand{\nonumberfootnote}[1]{%
  \renewcommand{\thefootnote}{}
  \footnotetext{\hspace*{-0.9em}#1}
  \renewcommand{\thefootnote}{\arabic{footnote}}
}

\nonumberfootnote{\textsuperscript{\Letter}\ Corresponding author: \email{xdliang328@gmail.com}.}

\input{sec/0_abstract}
\input{sec/1_intro}
\input{sec/2_related}
\input{sec/3_method}
\input{sec/4_exp}
\input{sec/5_conclusion}

\section*{Acknowledgements}
This work was supported in part by National Science and Technology Major Project (2020AAA0109704), National Science and Technology Ministry Youth Talent Funding No. 2022WRQB002, Guangdong Outstanding Youth Fund (Grant No. 2021B1515020061), Mobility Grant Award under Grant No. M-0461, Shenzhen Science and Technology Program (Grant No. GJHZ20220913142600001), Nansha Key RD Program under Grant No.2022ZD014.
\bibliographystyle{splncs04}
\bibliography{main}


\end{document}

%% file: sec/0_abstract.tex
\begin{abstract} 
Data-driven approaches for autonomous driving (AD) have been widely adopted in the past decade but are confronted with dataset bias and uninterpretability. 
Inspired by the knowledge-driven nature of human driving, recent approaches explore the potential of large language models (LLMs) to improve understanding and decision-making in traffic scenarios.
They find that the pretrain-finetune paradigm of LLMs on downstream data with the Chain-of-Thought (CoT) reasoning process can enhance explainability and scene understanding.
However, such a popular strategy proves to suffer from the notorious problems of misalignment between the crafted CoTs against the consequent decision-making, which remains untouched by previous LLM-based AD methods.
To address this problem, we motivate an end-to-end decision-making model based on multimodality-augmented LLM, which simultaneously executes CoT reasoning and carries out planning results.
Furthermore, we propose a reasoning-decision alignment constraint between the paired CoTs and planning results,
imposing the correspondence between reasoning and decision-making.
Moreover, we redesign the CoTs to enable the model to comprehend complex scenarios and enhance decision-making performance.
We dub our proposed large language planners with reasoning-decision alignment as \textit{\name{}}.
Experimental evaluations on the nuScenes and DriveLM-nuScenes benchmarks demonstrate the effectiveness of our \name{} in enhancing the performance of end-to-end AD systems. Specifically, our \name{} achieves state-of-the-art planning performance on the nuScenes dataset with 0.80 L2 error and 0.32 collision rate, and also achieves leading results on challenging DriveLM-nuScenes benchmarks with 0.82 L2 error and 0.38 collision rate.
\keywords{RDA-Driver \and LLMs \and Alignment \and  Autonomous Driving }
\end{abstract}

%% file: sec/1_intro.tex
\section{Introduction}
Autonomous driving has garnered significant attention and witnessed promising progress in recent years, with the potential to revolutionize industries such as transportation, logistics, and mobility services. 
Traditional approaches decouple the AD system into a stacked array of components responsible for task-specific problems including perception~\cite{liu2023bevfusion, liang2022bevfusion, li2022bevformer, huang2021bevdet}, prediction~\cite{gao2020vectornet, da2022path, gu2023vip3d}, and planning~\cite{gao2022cola, scheel2022urban, sadat2020perceive}, etc.
While effective for isolated applications, often they are criticized for the 
complexity of hand-crafting features and 
complex interactions among sub-modules.
Recent advances~\cite{hu2022st-p3, chen2024vadv2, hu2023uniad, jiang2023vad} attempt to streamline AD in end-to-end unified models which directly take as inputs the raw sensor data and generate the planning routes. Despite success, they raise challenges in terms of interpretability and robustness.

\begin{figure*}[t]
  \centering
    \includegraphics[width=1.0\linewidth]{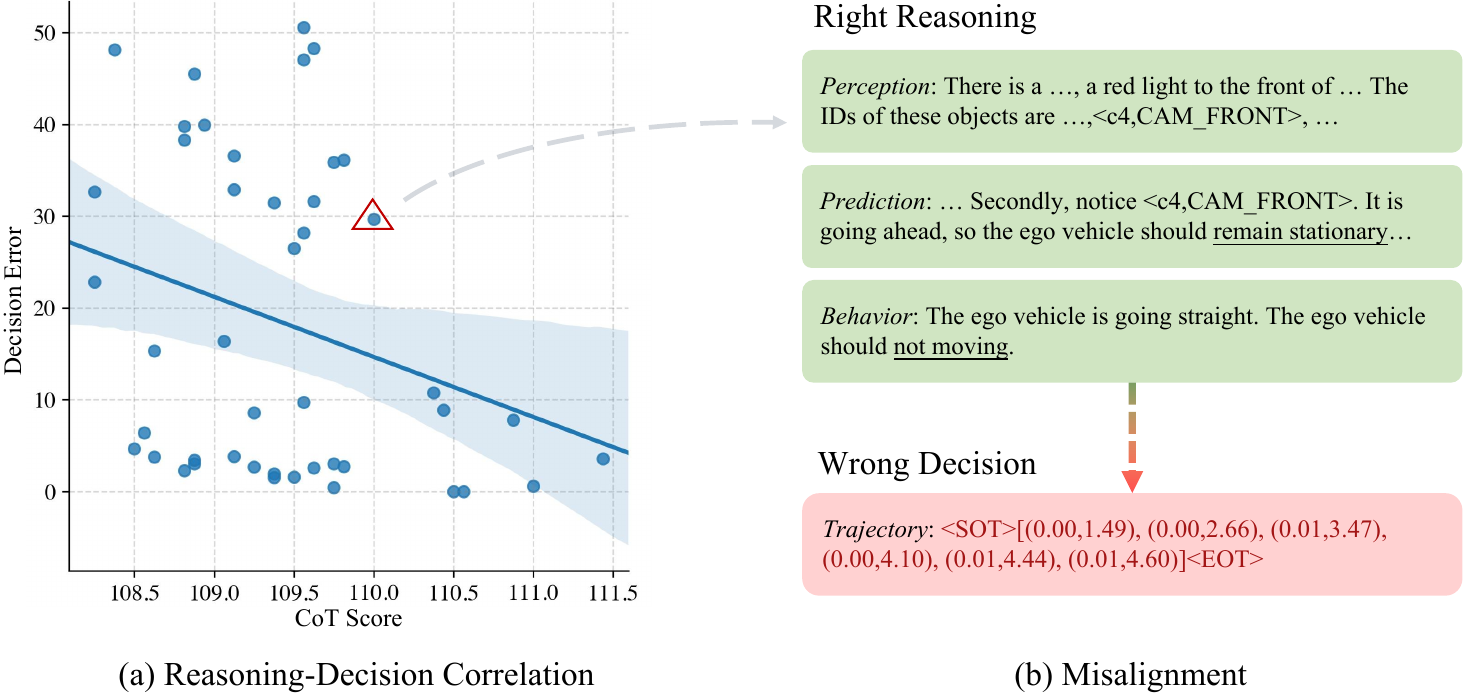}
  \caption{\textbf{Motivation of RDA-Driver.} (a) visualizes the distribution of the CoT score (higher is better) and the decision error of the predicted trajectory  (lower is better) of LLaVa~\cite{liu2024visual}, indicating the misalignment between the CoT reasoning the planning results. (b) shows an example of inconsistency between the CoT reasoning and the consequent decision. Although the model \textit{correctly} reasons the status of the current scene, i.e., noticing the front car and determining that it is not moving, the decision-making process follows \textit{wrong} plans for the ego vehicle to move forward.
 }
  \label{fig:motivation}
\end{figure*}
Large Language Models (LLMs), on the other hand, have shown significant potential in advancing models's capabilities of context understanding. Consequently, it is a temptation to equip the AD system with interpretability and generalization capabilities by means of LLMs.
A broad array of recent studies~\cite{pan2024vlp, learn_dmedriver,wang2023empowering,tian2024drivevlm,cui2023large,wang2023drivemlm,nie2023reason2drive,mao2023language,wang2023bevgpt,wen2023road,mao2023gptdriver,xu2023drivegpt4,wen2023dilu,cui2024drive,sha2023languagempc} have attempted to integrate vast world knowledge and robust logical reasoning abilities of LLMs into AD systems.
Typically, they adopt the pretraining-finetuning paradigm~\cite{liu2024visual} that fine-tunes LLMs on certain data with the chain-of-thought (CoT) reasoning process to enhance explainability and scene understanding.
More recently, there are growing concerns about the notorious problem regarding the misalignment~\cite{AFT, ramamurthy2022reinforcement, rafailov2024direct, zhao2023slic, yuan2023rrhf} of finetuned LLMs, i.e., the inconsistencies between assigned scores to CoTs and the accuracy of decision-making. A simple example is illustrated in~\cref{fig:motivation} (b). Here the model accurately reasons the front car ({\textsf{<c4, CAM\_FRONT>}}) and determines that it is not moving, while mistakenly asking the ego-vehicle to move forward on the final decision.  Surprisingly, this phenomenon has only recently been explored in~\cite{AFT, cobbe2021training, aggarwal2021explanations} and remains untouched by existing LLM-based AD methods, making it a timely problem.

To address this problem, 
we motivate an end-to-end decision-making model with multimodal LLMs, which transforms multi-view images into BEV feature representations as the input and can simultaneously carry out CoT reasoning and planning results.
Specifically, we propose the novel reasoning-decision alignment with a ranking loss between the CoT answers and the planning results, ensuring the model's explanation and its consequent conclusion are consistent and reliable.
Moreover, we redesign the CoT with logical thinking, so that the model could understand the scene and make inferences like a human driver, thereby making the entire decision-making process interpretable and interactive. 
Experimental evaluations on the nuScenes~\cite{caesar2020nuscenes} and DriveLM~\cite{sima2023drivelm} benchmarks demonstrate the effectiveness of our \name{} in achieving leading performance of end-to-end autonomous driving. 
Specifically, our \name{} achieves state-of-the-art end-to-end planning performance on the NuScenes dataset with 0.80 L2 error and 0.32 collision rates, and also achieves leading results on challenging DriveLM-nuScenes benchmarks with 0.82 L2 error and 0.38 collision rate.

In summary, our contributions can be summarized as follows:
\begin{itemize}
    \item We notice the misalignment issue in the current large language decision-making models, i.e., the inconsistencies between assigned scores to CoTs and the accuracy of decision-making.
    \item We propose a multimodal large language decision-making model with the reasoning-decision alignment, 
    \name{}, which improves the consistency of the model's explanation and conclusion. Moreover, we redesign the reasoning CoTs including perception, prediction, and decision-making, so that the model can complete highly interpretable decision-making tasks.
    \item Empirical evaluations on the NuScenes and DriveLM-nuScenes demonstrate our model can effectively improve decision-making performance and achieve flexible interactions and high interpretability.
\end{itemize}
\label{sec:intro}

%% file: sec/2_related.tex
\section{Related Work}
\label{sec:related}
\mypara{End-to-End Autonomous Driving.}
Modern autonomous driving solutions are broadly classified into two main categories: the conventional modular paradigm and the end-to-end approach. Recently, the end-to-end driving method has received extensive focus and research.
ST-P3~\cite{hu2022st-p3}, leveraging visual inputs, integrates feature learning across perception, prediction, and planning tasks, aiming for outputs that are more interpretable. This approach signifies a notable advancement in the field.
UniAD~\cite{hu2023uniad} introduces a systematic multi-task model, employing the transformer architecture to simultaneously model the interrelations among various tasks. This represents a significant stride in the development of versatile and comprehensive autonomous systems.
VAD~\cite{jiang2023vad} utilizes a method of vectorizing scene representations, thereby enhancing the computational efficiency of the model. This approach is advantageous for better understanding the surrounding environment and planning accurate and logical trajectories.

\mypara{Autonomous Driving with Large Language Models.}
With the remarkable capabilities demonstrated by large language models (LLMs) in interpretability and logical reasoning, a series of studies\cite{pan2024vlp, learn_dmedriver,wang2023empowering,tian2024drivevlm,cui2023large,wang2023drivemlm,nie2023reason2drive,mao2023language,wang2023bevgpt,wen2023road,mao2023gptdriver,xu2023drivegpt4,wen2023dilu,cui2024drive,sha2023languagempc} have been made to integrate LLMs into autonomous driving tasks. 
One approach involves constructing AD datasets related to text inputs and leveraging question-and-answer techniques for scene understanding and evaluation. 
For instance, LMDrive\cite{shao2023lmdrive} introduces a closed-loop dataset with navigation instructions using the CARLA\cite{dosovitskiy2017carla} simulator, while DriveLM\cite{sima2023drivelm} develops a dataset encompassing perception, prediction, and decision-making tasks based on nuScenes\cite{caesar2020nuscenes}. 
Another approach employs language inputs to further facilitate decision-making and planning tasks in AD using LLMs. 
GPT-Driver\cite{mao2023gptdriver}, for example, fine-tunes the GPT-3.5 to serve as a motion planner by converting detection and prediction outputs into text inputs. 
\cite{xu2023drivegpt4} constructs an interpretable end-to-end AD system with multimodal inputs. 
Leveraging the reasoning capabilities of LLMs, approaches like DME-Driver\cite{learn_dmedriver} and Reason2drive\cite{nie2023reason2drive} integrate logical reasoning with decision tasks, endowing models with human-like driving decision-making and reasoning abilities. However, these methods often overlook the issue of inconsistencies in causal relationships, which is crucial for safety-critical AD scenarios. 
In this paper, we propose a method that combines end-to-end AD with LLMs to enhance the consistency of reasoning and decision-making.

\mypara{Alignment in Large Language Models.}
Alignment of large language models aims to guide the model to exhibit thinking preferences similar to humans, rather than generating invalid inputs that contradict logical rules. 
 Existing alignment strategies can be broadly categorized into reinforcement learning and contrastive learning.
 LLMs aligned with Reinforcement Learning from Human Feedback (RLHF)\cite{ouyang2022training, ramamurthy2022reinforcement} have been widely used to compare different model generations.
 The solutions based on contrastive learning\cite{rafailov2024direct, zhao2023slic, yuan2023rrhf} do not require an explicit reward model by using a supervised fine-tuning paradigm, making them computationally efficient. \cite{AFT} enhances the reasoning capability of LLMs through contrastive loss function.
 However, previous alignment research has primarily focused on tasks such as math reasoning\cite{cobbe2021training} and commonsense reasoning\cite{aggarwal2021explanations}, neglecting the importance of reasoning and decision-making in AD. 
 As a motivator, this work is the first attempt to employ a contrastive learning to improve the alignment between logical reasoning and planning decisions in AD scenarios with high requirements for safety and timeliness.

%% file: sec/3_method.tex
\section{Method}
\label{sec: Method}
In this section, we introduce the proposed \name{} in detail. We first give an overview of the framework in \cref{subsec: overview}. Then, by considering the misalignment issue, we clarify the formulation of reasoning-decision alignment in \cref{subsec: alignment} and the redesigned CoT in \cref{subsec: CoT}.

\begin{figure*}[t]
  \centering
    \includegraphics[width=1.0\linewidth]{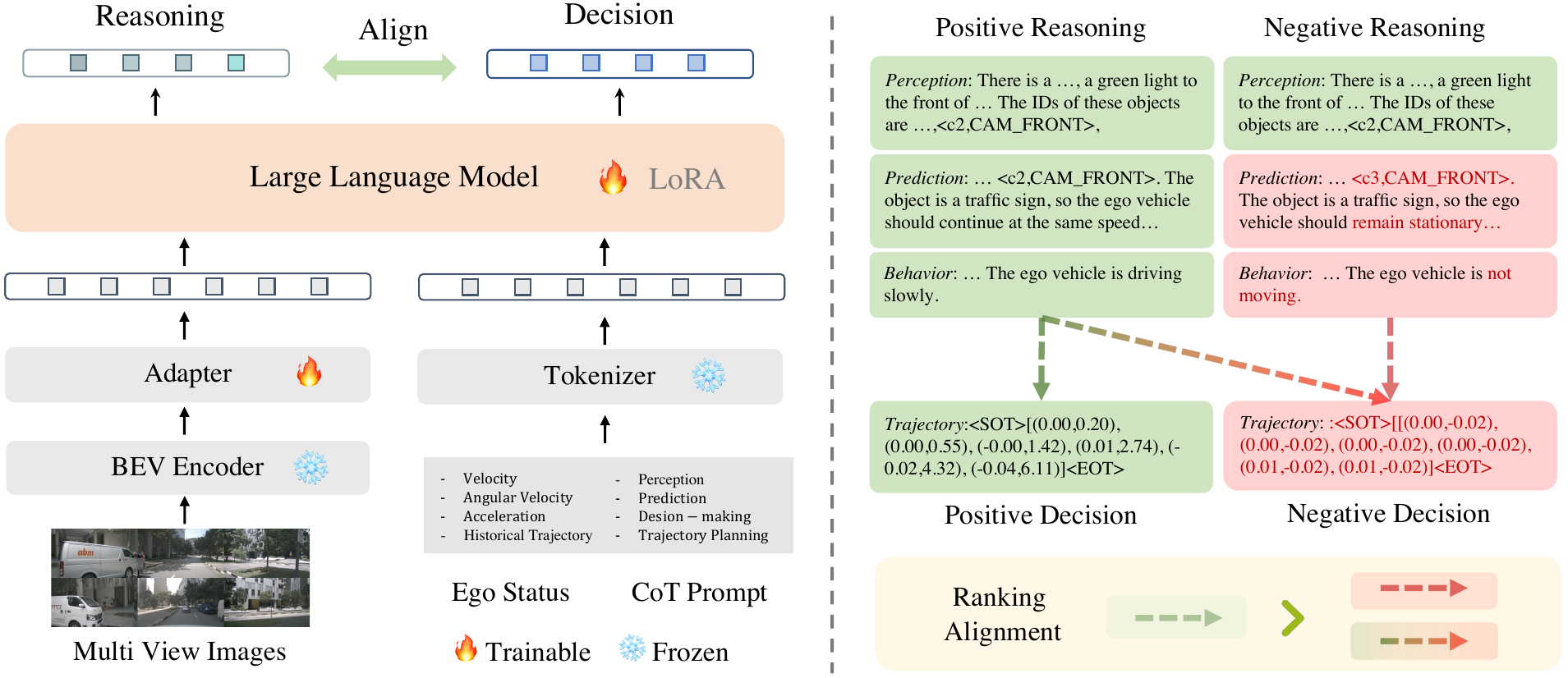}
  \caption{\textbf{Framework of \name{}.} \name{} takes the multi-view images, ego status, and multi-turn CoT prompt as input, and simultaneously carries out CoT reasoning and planning results. We construct multiple reasoning-decision samples with misalignment from both the vanilla fine-tuned model and similar scenarios. During training, we compute the token-average score as a measure of CoT answers. We utilize proposed contrastive loss to ensure the scores of positive samples are higher than those of generated negative samples.
  }
  \label{fig:framework}
\end{figure*}

\subsection{\name{} Overview}
\label{subsec: overview}
We propose a novel framework that utilizes the LLMs for AD system inference and decision-making.
Illustrated in \cref{fig:framework}, which bears resemblance to LLaVa~\cite{liu2024visual}, our proposed system, namely \name{}, comprises three core components: a vision encoder for processing multi-view images, an adapter module for translating visual representations into language-aligned visual tokens, and an LLM for receiving visual and language instruction tokens and generating responses. 
Specifically, for a given multi-view input $V$, we feed it into the vision encoder, followed by the adapter module, to obtain visual tokens.
Meanwhile, we tokenize the multi-round prompt texts $T$ to obtain the corresponding text tokens. 
Subsequently, all generated tokens are concatenated and passed the resultant inputs to the LLM to obtain the text output.
In this paper, we adopt Llama~\cite{touvron2023llama} as our LLM decoder. 
After generating the predicted labels, the de-tokenizer decodes them to restore human language.
\

\subsubsection{Vision Encoder.}
In large language models, the pre-trained CLIP~\cite{radford2021clip} model is widely used for encoding image features. 
However, the multiple perspectives of the images captured by our cameras differ significantly from the bird's-eye-view (BEV) perspectives needed to predict the trajectory. This perspective transformation is quite challenging for large language models. Inspired by the LSS~\cite{philion2020lift}, our vision encoder encodes the multi-perspective images into BEV features and ultimately transforms them into image tokens.
Formally, given $N$ multi-view images $V_{N}$, each with an extrinsic matrix $E_{k}$ and an intrinsic matrix $I_{k}$, the lift step projects each pixel from the 2D image onto the 3D voxel space based on its corresponding depth distribution while the splat step aggregates the feature values of pixels within each voxel using sum pooling.
We denote the BEV feature from the encoder as $B\in \mathbb{R}^{C \times H \times W }$, where $C$ represents the embedding dimension, and the $H$ and $W$ denote the height and width of the BEV.
To endow the BEV encoder with strong semantic features, we adopt object detection as a pre-training task. 
The object detection task requires predicting the significant objects in the environment.

\subsubsection{Adapter Module.}
The BEV features above are then passed into the BEV adapter module to be converted into visual tokens for the LLM.
Most existing methods utilize cross-attention operations such as Q-former~\cite{li2023blip} to obtain visual tokens. 
On the one hand, such operations may lose fine-grained details in the scene. 
On the other hand, training these parameters fully is challenging given the limited training samples (e.g. 4072 in our data). 
Similarly, directly performing pooling operations avoids the latter issue but sacrifices more visual information. 
Therefore, we propose a simple and efficient region encoding method.
For BEV features, we partition them into $h \times w$ grids. We then flatten each feature within a grid and consider it as a visual token. Consequently, we obtain visual input tokens $F_{V}\in \mathbb{R}^{(C*h*w) \times (H/h) \times (W/w) }$ by reducing the token number from $H \times W$ to $H/h \times W/w$.
Subsequently, we use a 2-layer MLP adapter to convert the tokens extracted to share the same dimension as the language token, which can then be fed into the LLM.


\subsection{Chain-of-Thought Alignment Tuning}
\label{subsec: alignment}
AD systems need to accurately understand changes in vehicle behavior and the surrounding environment, and respond appropriately. 
Causal CoT reasoning helps the system predict potential events by establishing causal relationships, thereby effectively planning driving routes and behaviors. 
For example, if an obstacle appears in front of the vehicle, the AD system must analyze and infer a series of events this obstacle may trigger, such as the need for emergency braking, lane change, reactions of surrounding vehicles, etc. 
These reasoning processes rely on an accurate understanding of causal relationships.
Any errors or inconsistencies may lead to incorrect decisions by the system, rendering the entire system unusable. 
Through CoT reasoning, AD systems can more accurately predict the possible consequences in various situations and adjust behaviors accordingly, thereby improving system safety and reliability. 
We need to ensure that the model's explanations and conclusions are aligned rather than contradictory during the reasoning process. 
Therefore, to ensure consistency in CoT reasoning, we adopt a consistency constraint loss to help the model understand the logical reasoning involved.

\subsubsection{Vanilla Chain-of-Though Fine-tuning.}
Given a training dataset consisting of N samples $\{V_{i}, T_{i}, A_{i}\}_{i=1}^{N}$, where $A$ denote the ground-truth trajectory for each prompt $T$. 
The multi-turn textual information $T$ encompasses the ego status information $T^{s}$ and CoT prompts $T^{c}$.
The ego status $T^{s}$ comprises information regarding the ego vehicle's velocity, heading angular velocity, acceleration, and the trajectory of the recent three frames.
We use the cross-entropy loss function to finetune the VLM with model parameter $\theta$:
\begin{equation}
        \mathcal{L}_{van} = -\sum_{j=1}^{|r_i|}{\log P(r_{i,j} | r_{i,<j}, T_{k,<j}, V; \theta)},
\label{eq:vanilla_loss}
\end{equation}
where $r_{i,j}$ is the $j$-th token of $r_i$, and $T_{k,<j}$ is the sub prompts of $T_i$ as we use multi-turn CoT.

\subsubsection{Dataset Generation.}
We first need to generate multiple multi-turn CoT for each sample in the training set.
We generate the dataset from both model and data perspectives simultaneously: \textbf{1. Model-base}: Firstly, we train the model $\theta$ using the vanilla loss function in Equation \ref{eq:vanilla_loss} to obtain a fine-tuned model $\widetilde{\theta}$. 
Then, we utilize $\widetilde{\theta}$ to sample $k$ generation results $\{ T_{i}, A_{i} \}_{i=1}^{k}$ for each input sample.
We rank these candidates based on the L2 metric between the predicted trajectory and the gt trajectory, resulting in a set of candidates of various qualities.
\textbf{2. Data-base}: Since there are samples in a driving process that exhibit similar scenes but completely different decisions, we therefore combine the annotated CoT labels from similar scenarios in the training set to form a set of positive and negative samples. 
For instance, given two samples ${(V_u, T_u^s, T_u^c, A_u), (V_v, T_v^s, T_v^c, A_v)}$ from the same scene but at different timestamps, we can permute and combine them to get multiple samples.
Regarding the sample associated with $V_u$, we can derive multiple negative samples ${(V_u, T_u^s, T_v^c, A_u)}, {(V_u, T_u^s, T_u^s, A_v)}, {(V_u, T_u^s, T_u^v, A_v)}$.
To align the behaviors of our model, we ensure that the quality of positive example ${(V_u, T_u^s, T_u^c, A_u)}$ is better than these negative examples.
The data generated from these two methods not only enhance their own decision-making and reasoning capabilities derived from the fine-tuned model but also learn the consistency of causal relationships from data with similar scenes but different reasoning and conclusions.

\subsubsection{Alignment Function.}
Inspired by the AFT\cite{AFT}, we align the scoring behaviors of LLMs are consistent with the golden standard assessment.
Firstly, We need to compute the token-average score for each sample:
\begin{equation}
        s_{\theta}^{c} = \frac{1}{|c|}\sum_{j=1}^{c}\log{P(c_j | c_{<j}, V, T; \theta)},
\label{eq:score_compute}
\end{equation}
where $V$ is the visual input and $T$ is the text input.
Subsequently, we guarantee that the scores assigned to positive CoTs surpass those assigned to negative CoT, ensuring consistency between the model's reasoning and its conclusions.
Specifically, for model-based generated data, we apply the ranking alignment loss to guide the model to generate more reliable CoTs and corresponding response:
\begin{equation}
        \mathcal{L}_{rank} = \log \{ 1 + \sum_{c_i \succ c_j }{\exp (\mathbf{D} ({s_{\theta}^{c_j}}) - s_{\theta}^{c_i} )} \},
\label{eq:loss_compute1}
\end{equation}
where we already rank the quality of all generated CoTs as a sequence $c_1 \succeq c_2 \succeq \cdots \succeq c_k$ base on the trajectory output, and $\mathbf{D}$ means detach operation.
Regarding the data-based generated data, we use the simple binary alignment loss to ensure that the scores of positive samples are higher than those of negative samples:
\begin{equation}
        \mathcal{L}_{binary} = \log \{ 1 + \sum_{c_n \in \mathbf{G_{N}} }{\exp (\mathbf{D} ({s_{\theta}^{c_n}}) - s_{\theta}^{c_p} )} \},
\label{eq:loss_compute2}
\end{equation}
where $\mathbf{G_{N}}$ denote the negative generated data, we also employ the detach operation $\mathbf{D}$.

Overall, the final training loss can be formulated as:
\begin{equation}
        \mathcal{L} = \mathcal{L}_{van} + \mathcal{L}_{rank} + \mathcal{L}_{binary}.
\label{eq:loss_all}
\end{equation}

\subsection{Driving with Chain-of-Thought Prompting}
\label{subsec: CoT}

\begin{wrapfigure}{r}{5.9cm}
\centering
\includegraphics[width=0.50\textwidth]{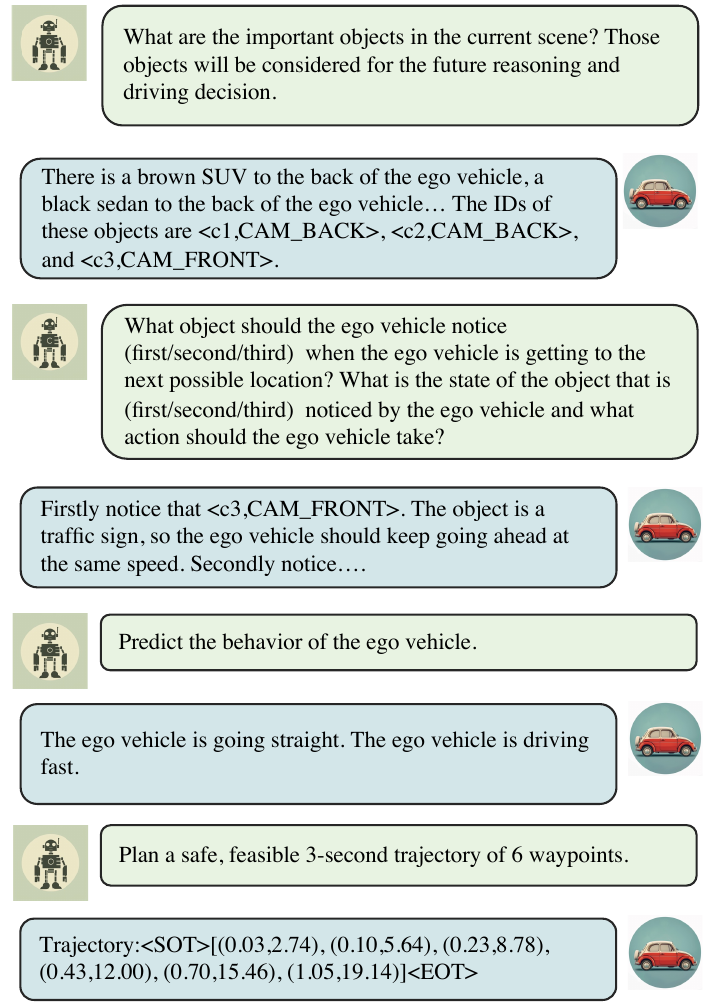}
\caption{\textbf{Illustrations of CoT prompt.}}
\label{fig: COT}
\end{wrapfigure}

In the human driving process, making driving decisions usually involves considering multiple factors. 
For instance, people typically first perceive the nearby key objects (such as cars, pedestrians, traffic lights, etc.), then determine the specific meanings of these objects or judge their driving trends, and finally deduce their own specific driving behaviors and plan the driving route. 
To endow our end-to-end system with similar interpretative capabilities, we introduce a novel reasoning module called the "Thinking Chain". 
We first mimic the human's first step of perception task to guide the model to identify the key objects 
, then continue to perform fine-grained motion prediction on these objects, distinguishing the relationship and priority of the influence of each object on the self-driving, and thus infer the driving behavior of the self-vehicle speed and direction to be consistent with the final planned route. 
We find that this strategy successfully combines the reasoning ability of the LLM with the background of AD, thus improving the accuracy of reasoning.
As illustrated in \cref{fig: COT}, our overall logical chain includes the following steps:


\noindent \textbf{\textit{Perception}}: Identifying key objects and their positions.
During road driving, humans instinctively identify key aspects of the scene, focusing on elements that directly affect our driving. 
These may include traffic signals or other instances that interact with the ego vehicle. Identifying key objects in the scene is fundamental to understanding the driving environment and planning driving behavior. 
Thus, we use the prompt \emph{"What are the important objects in the current scene? Those objects will be considered for the future reasoning and driving decision."} for the perception part, asking the system to identify the important objects in the scene and their locations.

\setlength\intextsep{0pt}
\setlength\columnsep{10pt}

\noindent \textbf{\textit{{Prediction}}}: Estimating the motion and intentions of key objects.
How to handle interactions with key objects is a critical aspect of human understanding of the scene. 
Human drivers focus on elements that may affect driving decisions and their relationships, logically describing the specific meanings of these objects and prioritizing instances. 
For example, in a traffic light scenario, different colors are determinant factors for driving behavior, and pedestrians ahead are more worth paying attention to than parallel vehicles. 
We select the prompt whcih requires the system to accurately analyze driving interactions. The prompts are as follows: \emph{"What object should the ego vehicle notice (first/second/third) when the ego vehicle is getting to the next possible location? What is the state of the object that is (first/second/third) noticed by the ego vehicle, and what action should the ego vehicle take?"}.
Logical descriptions can guide the model to make correct judgments and avoid hidden risks.

\noindent \textbf{\textit{Decision-making}}: Determining the appropriate driving actions based on the perceived environment and predicted future states.
The decision-making process of human driving is highly logical and inductive, involving the driver's ability to consider various factors and come up with appropriate actions. 
Therefore, we select a simple prompt \emph{"Predict the behavior of the ego vehicle."} to allow the autonomous driving system to summarize the perception and prediction results and complete the potential thinking process. 
The system needs to answer the direction in which the ego vehicle should drive (e.g., forward or turn) and the speed of the ego vehicle (e.g., stop or accelerate). 
This logical ability to mimic human decision-making makes the system's judgments more reliable.

\noindent \textbf{\textit{Planning}}: Forecasting waypoints of ego vehicle's future trajectory.
The final output of a complete driving system should be directly output control signals or trajectories. 
Therefore, while ensuring the correctness of natural language logical reasoning, we need to finally output a future driving trajectory using the prompt \emph{"Plan a safe, feasible 3-second trajectory of 6 waypoints."}.
Trajectory planning is ultimately what the system needs to complete and is a crucial part of evaluating the model's capabilities.


Certainly, these comprehensive CoT prompts consider the perception, prediction, and decision-making methods in end-to-end AD.

%% file: sec/4_exp.tex
\section{Experiment}
\label{sec:exp}
In this section, we introduce the experimental setup in \cref{subsec: setup} and compare with leading approaches in \cref{subsec: main_res}. 
The analysis is presented in \cref{subsec: ablations}. 

\subsection{Experimental Setting}
\label{subsec: setup}
\mypara{Datasets.}
We use nuScenes~\cite{caesar2020nuscenes} to evaluate the planning task, which is a challenging and popular benchmark in the AD. The dataset is a multi-sensor dataset with 1,000 scenes and each scene lasts for 20 seconds. There are 28,130 training samples and 6,019 validation samples.
To alleviate the "ego status" problem in the open-loop evaluation, we also conduct experiments on DriveLM-nuScenes~\cite{sima2023drivelm} dataset, which is built upon the nuScenes to offer comprehensive and accurate question-answer pairs (QAs). 
These QAs cover various aspects of the driving process including perception, prediction, and planning, thus providing a thorough understanding of AD scenarios.
There are 4,072 training samples and 799 validation samples in DriveLM-nuScenes.

\mypara{Evaluation Metrics.}
We follow the common practice from previous works\cite{hu2022st-p3, hu2023uniad}, employing two metrics to assess the quality of output trajectories: L2 error (in meters) and collision rate (in percentage). The L2 error measures the similarity between the predicted trajectories and actual human driving trajectories. To determine the frequency of collisions between the ego vehicle and other objects, we simulate the ego vehicle's trajectory by placing a bounding box at each waypoint and check for collisions with other oriented bounding boxes of objects detected in the scene. Similar to most trajectory prediction works in nuScenes, we evaluate the motion planning results within a 3-second timeframe, and assess the quality of the output trajectories for 1s, 2s, and 3s time horizons.
Moreover, following GPT-Driver\cite{mao2023gptdriver}, we implement two evaluation methods (ST-P3 metric and UniAD metric) for a fair comparison with different models.

\mypara{Implementation Details.}
We adopt LLaVa\cite{liu2024visual} as the competitive baseline.
We use the experimental setting of camera-based 3D detection from BEVFusion~\cite{liang2022bevfusion} to pre-train the BEV encoder. Subsequently, during the fine-tuning process of the large models, we do not update the parameters of the BEV encoder.
\name{} is trained for 10 epochs with LoRA fine-tuning strategy. 
We use 8 NVIDIA A100 GPUs with 4 samples per each GPU for training.
Also we use AdamW as the optimizer and cosine annealing scheduler as the learning rate scheduler with an initial learning rate of 1$e$-4.
As we rely on CoT labels as supervision during fine-tuning, we use only 4072 samples from the DriveLM-nuScene dataset for training, which accounts for approximately 1/7 of the nuScenes training set.
We conduct evaluations on both the complete nuScenes validation set and DriveLM-nuScene validation set to demonstrate the effectiveness of the models.

\subsection{Main Results}
\label{subsec: main_res}
\input{latex/table/main_result}
We compare the \name{} with current state-of-art methods and report the result in two benchmraks, nuScenes and DriveLM-nuScenes.
We list each model's average performance and group them by different evaluation methods.
As illustrated in \cref{tab:sota-plan}, \name{} outperforms or is comparable to the prior works in both L2 and collision rate across the two different evaluation methods.
It is worth noting that our collision rate within 1 second is 0.
These excellent results demonstrates the effectiveness of our approach in generating human-like driving trajectories and its capability to plan safe driving paths.
It is worth noting that an increasing number of learning-based end-to-end AD models have demonstrated excellent performance in planning, such as UniAD\cite{hu2023uniad} and VAD\cite{jiang2023vad}. 
However, by fine-tuning large language models on a small amount of CoT annotated data (only 1/7 of their training size), our approach achieves comparable or even better performance. 
This indicates our method's ability to effectively leverage the knowledge embedded within LLMs for planning tasks, offering a potential solution for end-to-end AD. 
Furthermore, compared to models like GPT-Driver\cite{mao2023gptdriver}, which leverage the unique capabilities of ChatGPT but require integrating results from other foundational models such as perception and prediction, our method achieves superior performance in both metrics of the two evaluation methods without relying on these priors. 
Models like DME-Driver\cite{learn_dmedriver} are structurally similar to ours, which also feed multimodal information into LLMs. 
By constructing reasoning-decision alignment loss functions, our method achieves low collision rates and significantly improves the L2 metric, demonstrating the importance of aligned reasoning and decision-making for AD.

Recently, some works~\cite{li2023ego, zhai2023rethinking} have mentioned the problem of trajectory prediction overly relying on ego status inputs in open-loop evaluation. 
To address this issue, we conduct experiments on a more challenging dataset, DriveLM-nuScenes. 
This dataset is curated by selecting key frames from the complete nuScenes data, where the ego vehicle's intention changes, and the ego history is not strongly indicative of future behavior or motion. 
This helps alleviate the "ego status" problem. 
In this case, to further demonstrate the effectiveness of our approach, we perform experiments on this dataset and compare it with existing literature.
As shown in \cref{tab:sota-plan2}, learning-based end-to-end models, UniAD and ST-P3, perform poorly in this setting. 
For instance, the L2 metric for the UniAD model increases from 1.03 to 3.00. GPT-Driver also shows significant changes in collision rate from 0.44 to 0.67. 
However, our model exhibits only minor fluctuations  ($0.01\sim0.06$) and continues to maintain excellent performance.



\subsection{Ablation Study}
\subsubsection{Effectiveness of Alignment Loss.} 

\label{subsec: ablations}

\setlength\intextsep{0pt}
\setlength\columnsep{10pt}
\begin{wraptable}{r}{0.4\textwidth}
    \centering
    \setlength\tabcolsep{2.6pt}
    \begin{tabularx}{0.35\textwidth}{c|c}
        \toprule
        Alignment loss & CoT score \\
        \toprule
        \xmark  &  62 \\
        \rowcolor{mygray} \cmark  & \bf 71 \\
        \bottomrule
    \end{tabularx}
    \caption{Alignment loss enhances the effectiveness of CoT reasoning.}
    \label{tab:ablation_align}
\end{wraptable}

We conduct thorough experiments, including four different loss function to validate the individual effectiveness of the proposed alignment objective in~\cref{tab:ablation_loss}. 
Generally, we utilize the ranking loss to compare the multiple samples generated by the fine-tuned model, aiding the model in implicit logical reasoning and enhancing its understanding of each task. 
Additionally, using binary loss in training similar samples serves the purpose of assisting the model in distinguishing between different scenarios and better understanding subtle differences within the scene. 
It helps the model capture the similarities and differences between samples, enhancing its understanding of complex situations and promoting the development of its logical reasoning capabilities.
To further validate the effectiveness of the alignment loss function in enhancing CoT logical reasoning and consequently producing accurate planning decision, we conduct CoT metric measurements using ChatGPT\cite{achiam2023gpt}. 
It is noteworthy that since the Drivelm-nuScenes validation set does not disclose its CoT labels, we partition the training set and utilize 200 training samples to evaluate the quality of the model's CoT responses. 
By providing both the predicted CoTs and GT as inputs, we employ the same prompt used in DriveLM\cite{sima2023drivelm} to prompt GPT3.5 to score the predicted results on a scale from 0 to 100. 
For each sample, we evaluate each round of answers separately and compute the average as the CoT response score.
The final CoT score of the model is represented as:
\begin{equation}
        s_{cot} = \frac{1}{|c|}\sum_{i=1}^{c}(\frac{1}{|k|}\sum_{j=1}^{k} {\rm GPT3.5} (r_i^{j}, g_{i}^{j})), 
\label{eq:score_chatgpt}
\end{equation}

where $r_i$ is the $i$-th CoT response predictions and $g_i$ is the $i$-th corresponding GT, and we have $k$ rounds.
As illustrated in \cref{tab:ablation_align}, by introducing the alignment loss, our CoT score also improved from 62 to 71, demonstrating that our approach not only enhances reasoning ability but also improves the final decision-making performance through correct logical inference.

\begin{table*}[t]
\centering
\addtolength{\tabcolsep}{4.7pt}
 \begin{tabularx}{0.91\textwidth}{cc|ccc|c|ccc|c}
\toprule
\multirow{2}{*}{ranking} & \multirow{2}{*}{binary} & 
\multicolumn{4}{c|}{L2 (m) $\downarrow$} & 
\multicolumn{4}{c}{Collision (\%) $\downarrow$} \\
\cmidrule(){3-10} &
  & 1s & 2s & 3s & Avg. & 1s & 2s & 3s & Avg. \\
\toprule

\xmarkg  & \xmarkg  & 0.25 & 0.76 & 1.55 & 0.85 & 0.00 & 0.27 & 1.06 & 0.44\\

\colorbox{mycyan}{\cmark}  & \xmarkg  &  0.23 & 0.73 & 1.53 & 0.83 & 0.00 & 0.17 & 0.86 & 0.34\\

\xmarkg  & \colorbox{mycyan}{\cmark}  & 0.25 & 0.77 & 1.59 & 0.87 & 0.02 & 0.15 & 1.00 & 0.39 \\ 

\midrule 
\rowcolor{mygray} \colorbox{mycyan}{\cmark}  & \colorbox{mycyan}{\cmark}  & 0.23 & 0.73 & 1.54 & \textbf{0.80} & 0.00 & 0.13 & 0.83 & \textbf{0.32} \\

\bottomrule
\end{tabularx} 
\caption{Ablation for our reasoning-decision alignment loss.}
\label{tab:ablation_loss}
\end{table*}

\begin{table*}[t]
       \centering
       \setlength\tabcolsep{4.3pt}
       \begin{tabularx}{0.91\textwidth}{c|ccc|c|ccc|c}
       \toprule
       \multirow{2}{*}{Visual encoder} &\multicolumn{4}{c|}{L2 (m) $\downarrow$} & \multicolumn{4}{c}{Collision (\%) $\downarrow$} \\
       \cmidrule(){2-9}
       & 1s & 2s & 3s & Avg. & 1s & 2s & 3s & Avg. \\
       \toprule
       CLIP-based  & 0.25 & 0.75 & 1.54 & 0.84 & 0.13 & 0.33 & 0.96 & 0.48 \\ 
       BEV-based + Pooling & 0.33 & 1.36 & 2.35 & 1.35 & 0.13 & 0.45 & 2.03 & 0.87 \\
    \rowcolor{mygray}   BEV-based + Flattening & 0.25 & 0.76 & 1.52 & \textbf{0.84} & 0.00 & 0.27 & 0.96 & \textbf{0.41} \\
       \bottomrule
       \end{tabularx}
   \caption{The effect on the visual encoder and visual tokens.}
   \label{tab:ablation_chatgpt_encoder}
\end{table*}

\subsubsection{Comparison between Visual Encoders.} 
To demonstrate the introduced encoder, we compare the CLIP-based encoder and BEV-based encoder in \cref{tab:ablation_chatgpt_encoder}.
The pre-trained CLIP model\cite{radford2021clip} is commonly used by LLMs for various visual tasks, while the BEV encoder is a common visual encoder in AD models. 
Therefore, we compare the effectiveness of these two visual encoders in the context of an end-to-end AD LLM.
From the experimental results, it is evident that simply replacing the CLIP encoder with the BEV encoder and obtaining visual input tokens through pooling operations has adverse effects on the final decision-making. However, partitioning the BEV features into regions and flattening them yields better results than the CLIP model.
On one hand, in AD scenarios, there exists a disparity between the input images and the output trajectories, making the representation of BEV features more suitable. 
On the other hand, preserving the BEV features when converting them into input tokens for the LLMs is crucial, particularly for tasks requiring fine-grained trajectory prediction regression.

\input{latex/table/ablation}

\subsubsection{Effectiveness of chain-of-Thought Redesign.} 
To demonstrate the introduced in redesigned CoTs, we compare through experiments with multiple sets of CoT combinations in \cref{tab:ablation_cot}.
We complete experiments from two aspects: the CoT category and the number of QA rounds. 
First, comparing the results between 1 and 2, as well as between 2 and 6, we observe that multi-round CoTs contributes to the model's ability to perform logical reasoning for complex decision tasks, resulting in more accurate predictions. 
Second, comparing the experiments in 3, 4, 5 with 6, it is evident that holistic logical reasoning considering perception, prediction, and decision-making is crucial for accurate trajectory prediction. 
Any missing component weakens the reasoning capability.


%% file: latex/table/main_result.tex
{
\begin{table*}[t]
\centering
\addtolength{\tabcolsep}{2.7pt}
 \begin{tabularx}{0.91\textwidth}{l|ccc|c|ccc|c}
\toprule

\multirow{2}{*}{Method} &
\multicolumn{4}{c|}{L2 (m) $\downarrow$} & 
\multicolumn{4}{c}{Collision (\%) $\downarrow$} \\
& 1s & 2s & 3s & Avg. & 1s & 2s & 3s & Avg. \\

\toprule
\multicolumn{3}{l}{\textit{\textbf{ST-P3 metrics}}} \\
\midrule
ST-P3~\cite{hu2022st-p3} & 1.33 & 2.11 & 2.90 & 2.11 & 0.23 & 0.62 & 1.27 & 0.71 \\
VAD~\cite{jiang2023vad} & 0.17 & \bf 0.34 & \bf 0.60 & \textbf{0.37} & 0.07 & 0.10 & \bf 0.24 & 0.14 \\
GPT-Driver~\cite{mao2023gptdriver} & 0.20 & 0.40 & 0.70 & 0.44 & 0.04 & 0.12  & 0.36 & 0.17 \\
DriveVLM~\cite{tian2024drivevlm} & 0.18 &0.34& 0.68 &0.40 &0.10 &0.22 &0.45 &0.27 \\
\midrule
\rowcolor{mygray} \textbf{\name{} (ours)} & \bf 0.17 & 0.37 & 0.69 & 0.40 & \bf 0.01 & \bf 0.05  & 0.26 & \textbf{0.10} \\
\midrule
\multicolumn{3}{l}{\textit{\textbf{UniAD metrics}}} \\
\midrule
NMP~\cite{learn_nmp} & - & - & 2.31 & - & - & - & 1.92 & - \\
SA-NMP~\cite{learn_nmp} & - & - & 2.05 & - & - & - & 1.59 & - \\
FF~\cite{learn_ff} & 0.55 & 1.20 & 2.54 & 1.43 & 0.06 & 0.17 & 1.07 & 0.43 \\
EO~\cite{learn_eo} & 0.67 & 1.36 & 2.78 & 1.60 & 0.04 & 0.09 & 0.88 & 0.33 \\

UniAD~\cite{hu2023uniad} & 0.48 & 0.96 & 1.65 & 1.03 & 0.05 & 0.17 & 0.71 & 0.31 \\
GPT-Driver~\cite{mao2023gptdriver} & 0.27  & 0.74 & \bf 1.52 & 0.84 & 0.07 & 0.15 & 1.10 & 0.44 \\

DME-Driver\cite{learn_dmedriver} & 0.45 & 0.91 & 1.58 & 0.98 & 0.05 & 0.15 & \bf 0.68 & \textbf{0.29} \\
\midrule
\rowcolor{mygray} \textbf{\name{} (ours)} & \bf 0.23  & \bf 0.73 & 1.54 & \textbf{0.80} & \bf \textcolor{purple}{0.00} & \bf 0.13 & 0.83 & 0.32 \\
\bottomrule
\end{tabularx}
\caption{Motion planning performance on nuScenes benchmark. Our approach significantly outperforms or is comparable to the prior works with a small number of labels.
}
\label{tab:sota-plan}
\end{table*}
}

{
\begin{table*}[!t]
\centering
\addtolength{\tabcolsep}{2.7pt}
 \begin{tabularx}{0.91\textwidth}{l|ccc|c|ccc|c}
\toprule
\multirow{2}{*}{Method} &
\multicolumn{4}{c|}{L2 (m) $\downarrow$} & 
\multicolumn{4}{c}{Collision (\%) $\downarrow$} \\
 & 1s & 2s & 3s & Avg. & 1s & 2s & 3s & Avg. \\
\toprule
\multicolumn{3}{l}{\textit{\textbf{ST-P3 metrics}}} \\
\midrule
ST-P3~\cite{hu2022st-p3} & 1.28 & 2.03 & 2.81 & 2.04 & 0.14 & 0.72 & 1.28 & 0.71 \\
GPT-Driver~\cite{mao2023gptdriver} & 0.22 & 0.43 & 0.73 & 0.46 & 0.00 & 0.13  & 0.46 & 0.19 \\
\midrule
\rowcolor{mygray}  \textbf{\name{} (ours)} &\bf 0.18 & \bf0.38 &\bf 0.68 & \textbf{0.41} & \bf 0.00 &\bf 0.06  & \bf 0.36 & \textbf{0.14} \\
\midrule
\multicolumn{3}{l}{\textit{\textbf{UniAD metrics}}} \\
\midrule
UniAD~\cite{hu2023uniad} & 0.47 & 1.80 & 3.73 & 3.00 & 0.13 & 0.53 & 1.50 & 0.72 \\
GPT-Driver~\cite{mao2023gptdriver} & 0.30  & 0.77 & 1.54 & 0.87 & 0.00 & 0.38 & 1.63 & 0.67 \\
\midrule
\rowcolor{mygray} \textbf{\name{} (ours)} & \bf 0.25  & \bf 0.72 & \bf 1.49 & \textbf{0.82} & \bf 0.00 & \bf 0.13 & \bf 1.00 & \textbf{0.38} \\
\bottomrule
\end{tabularx}
\caption{Motion planning performance in DriveLM-nuScenes validation set. Ours maintain excellent performance in terms of L2 and collision rate.
}
\label{tab:sota-plan2}
\end{table*}
}

%% file: latex/table/ablation.tex
\begin{table*}[t]
\centering
\addtolength{\tabcolsep}{3.2pt}
 \begin{tabularx}{0.91\textwidth}{c|ccc|ccc|c|ccc|c}
\toprule
\multirow{2}{*}{ID} &\multirow{2}{*}{PER} & \multirow{2}{*}{PRE} & \multirow{2}{*}{DM} &

\multicolumn{4}{c|}{L2 (m) $\downarrow$} & 
\multicolumn{4}{c}{Collision (\%) $\downarrow$} \\
\cmidrule(){5-12} &
 & & & 1s & 2s & 3s & Avg. & 1s & 2s & 3s & Avg. \\
\midrule
\multicolumn{11}{l}{\textit{\textbf{Single-Turn}}}\\
\midrule

1 &\xmarkg & \xmarkg & \xmarkg    & 0.25 & 0.88 & 1.90 & 1.01 & 0.00 & 0.38 & 1.06 & 0.48\\

2 & \colorbox{mycyan}{\cmark} & \colorbox{mycyan}{\cmark} & \colorbox{mycyan}{\cmark}   &
0.23 & 0.78 & 1.68 & \textbf{0.90} & 0.00 & 0.18 & 1.18 & \textbf{0.45}\\

\midrule
\multicolumn{11}{l}{\textit{\textbf{Multi-Turn}}}\\
\midrule
3 & \colorbox{mycyan}{\cmark} & \xmarkg & \xmarkg   &  0.23 & 0.79 & 1.58 & 0.87 & 0.02 & 0.22 & 0.88 & 0.37\\

4 &{\cmarkg} & \colorbox{mycyan}{\cmark} & \xmarkg   & 0.24 & 0.79 & 1.66 & 0.89 & 0.00 & 0.13 & 1.01 & 0.38 \\

5 &\xmarkg & \xmarkg & \colorbox{mycyan}{\cmark}   &  0.23 & 0.79 & 1.72 & 0.91 & 0.00 & 0.18 & 1.05 & 0.41 \\

\midrule
\rowcolor{mygray}  6 & \colorbox{mycyan}{\cmark} & \colorbox{mycyan}{\cmark}& 
\colorbox{mycyan}{\cmark}   & 0.23 & 0.73 & 1.54 & \textbf{0.80} & 0.00 & 0.13 & 0.83 & \textbf{0.32} \\

\bottomrule
\end{tabularx} 
\caption{Ablation for the redesigned CoTs.}
\label{tab:ablation_cot}
\end{table*}

%% file: sec/5_conclusion.tex
\section{Conclusion}
This work presents a LLM-based AD model RDA-Driver with reasoning-decision alignment, which offers better interpretability and robustness. To our knowledge, this work first  identifies the problem of misalignment residing in the LLM-based AD method and is the first attempt to solve this problem. To this end, we propose the reasoning-decision alignment which imposes constraints between the CoT process and the subsequent planning results by contrastive learning. Furthermore, we design multi-turn CoT which encourages the model to think about the driving scenarios holistically. 
The input modality is limited to multi-view images and we may explore video input in future work.
